\documentclass[lettersize,journal]{IEEEtran}
\usepackage{amsmath,amsfonts}
\usepackage{algorithmic}
\usepackage{algorithm}
\usepackage{array}
\usepackage[caption=false,font=normalsize,labelfont=sf,textfont=sf]{subfig}
\usepackage{textcomp}
\usepackage{stfloats}
\usepackage{url}
\usepackage{verbatim}
\usepackage{graphicx}
\usepackage{cite}
\usepackage{booktabs}
\usepackage{multirow}
\usepackage{tabularx}
\usepackage{makecell}
\usepackage{soul}
\usepackage[table, xcdraw]{xcolor}
\usepackage[hidelinks]{hyperref}

\begin{document}

\title{EDTformer: An Efficient Decoder Transformer for Visual Place Recognition}

\author{Tong Jin, Feng Lu, Shuyu Hu, Chun Yuan,~\IEEEmembership{Senior Member,~IEEE,} Yunpeng Liu,~\IEEEmembership{Member,~IEEE}
\thanks{Tong Jin and Shuyu Hu are with the Key Laboratory of Opto-Electronic Information Processing, Chinese Academy of Sciences, Shenyang 110016, China, and the Shenyang Institute of Automation, Chinese Academy of Sciences, Shenyang 110016, China, and also with the School of Computer Science and Technology, University of Chinese Academy of Sciences, Beijing 100049, China (e-mail: jintong@sia.cn; hushuyu@sia.cn).}
\thanks{Yunpeng Liu is with the Shenyang Institute of Automation, Chinese Academy of Sciences, Shenyang 110016, China (e-mail: ypliu@sia.cn). Yunpeng Liu is the Corresponding Author.}
\thanks{Feng Lu and Chun Yuan are with Tsinghua Shenzhen International Graduate School, Tsinghua University, Shenzhen, China (e-mail: 
lf22@mails.tsinghua.edu.cn; yuanc@sz.tsinghua.edu.cn).}}

\markboth{Journal of \LaTeX\ Class Files,~Vol.~14, No.~8, August~2021}%
{Shell \MakeLowercase{\textit{et al.}}: A Sample Article Using IEEEtran.cls for IEEE Journals}

\IEEEpubid{\begin{minipage}{\textwidth}\ \centering
		Copyright \copyright 2025 IEEE. Personal use of this material is permitted. \\
		However, permission to use this material for any other purposes must be obtained 
		from the IEEE by sending an email to pubs-permissions@ieee.org.
\end{minipage}}

\maketitle

\begin{abstract}
Visual place recognition (VPR) aims to determine the general geographical location of a query image by retrieving visually similar images from a large geo-tagged database. To obtain a global representation for each place image, most approaches typically focus on the aggregation of deep features extracted from a backbone through using current prominent architectures (e.g., CNNs, MLPs, pooling layer, and transformer encoder), giving little attention to the transformer decoder. However, we argue that its strong capability to capture contextual dependencies and generate accurate features holds considerable potential for the VPR task. To this end, we propose an Efficient Decoder Transformer (EDTformer) for feature aggregation, which consists of several stacked simplified decoder blocks followed by two linear layers to directly produce robust and discriminative global representations. Specifically, we do this by formulating deep features as the keys and values, as well as a set of learnable parameters as the queries. Our EDTformer can fully utilize the contextual information within deep features, then gradually decode and aggregate the effective features into the learnable queries to output the global representations. Moreover, to provide more powerful deep features for EDTformer and further facilitate the robustness, we use the foundation model DINOv2 as the backbone and propose a Low-rank Parallel Adaptation (LoPA) method to enhance its performance in VPR, which can refine the intermediate features of the backbone progressively in a memory- and parameter-efficient way. As a result, our method not only outperforms single-stage VPR methods on multiple benchmark datasets, but also outperforms two-stage VPR methods which add a re-ranking with considerable cost. Code will be available at \url{https://github.com/Tong-Jin01/EDTformer}.
\end{abstract}


\begin{figure}[t]
    \centering
    \includegraphics[width=1.0\columnwidth]{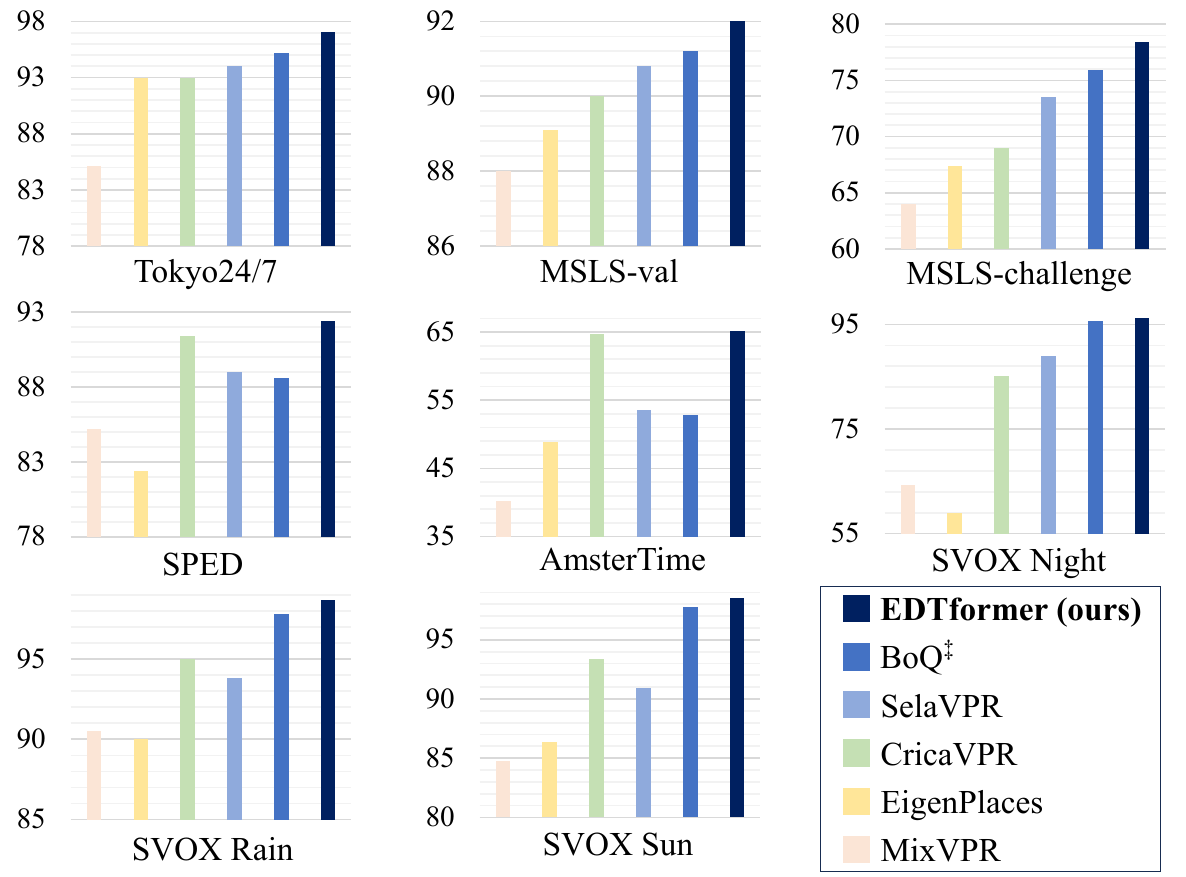}
    \caption{
    The performance comparison (Recall@1) on multiple benchmark datasets between our method and current state-of-the-art VPR methods, such as MixVPR \cite{mixvpr}, EngenPlaces \cite{eigenplaces}, CricaVPR \cite{cricavpr}, SelaVPR \cite{selavpr} and BoQ \cite{BoQ}. $\ddag$ We reproduce the results of BoQ by strictly following its training pipeline, except for keeping the same image size for both training ($224\times224$) and inference ($322\times322$) as our method. Our EDTformer consistently shows obvious advantages over other methods in diverse VPR scenarios, including viewpoint variations and condition changes (MSLS \cite{msls}), severe lighting changes (Tokyo24/7 \cite{tokyo247}), various low-quality and high-scene-depth place images (SPED \cite{sped}), place image modality changes (AmsterTime \cite{amstertime}) and varying weather conditions (SVOX \cite{svox}).
    }
    \label{fig:R@1 Performance}
\end{figure}

\section{Introduction}
\IEEEPARstart{V}{isual} place recognition, also referred to as visual geo-localization \cite{benchmark}, plays an essential role in autonomous driving \cite{bresson2017simultaneous}, mobile robot localization \cite{xu2020probabilistic,tcsvtrobot}, augmented reality \cite{tcsvtvpr2}, etc. Therefore, it has attracted considerable interest in the fields of computer vision and robotics over the past decade. However, there still exist various challenges that we have to face in VPR, such as environment variations, viewpoint changes and perceptual aliasing \cite{survey,Tutorial}. Addressing these issues while achieving a good accuracy-efficiency trade-off is very difficult but valuable, particularly for single-stage VPR methods that only employ global features. 

\IEEEpubidadjcol

VPR is generally addressed as a special image retrieval task \cite{cao2020unifying}. Each place image is represented by a global feature, and then the nearest neighbor search is performed in the feature space to obtain the best-matched images of the query. The global features are typically derived from the aggregation of deep features, utilizing some techniques such as NetVLAD \cite{netvlad}, GeM \cite{GeM} pooling or their variants \cite{patch-netvlad,tcsvtvpr1,Cosplace}. Unfortunately, such global features usually can not perform well in challenging scenes. To get robust global features and thereby address the main challenges in VPR, recent single-stage VPR research has focused on exploring novel methods to achieve feature aggregation using some popular architectures. For instance, MixVPR \cite{mixvpr} incorporates global relationships into each feature map through a stack of Feature-Mixer, which just consists of the multi-layer perceptrons (MLPs). CricaVPR \cite{cricavpr} proposes a cross-image encoder to calculate the correlation between representations of multiple images in the same batch for robust global features. SALAD \cite{SALAD} redefines the soft assignment of local features in NetVLAD as an optimal transport problem, in which both CNNs and MLPs are fully utilized. BoQ \cite{BoQ} attempts to aggregate features by utilizing additional transformer encoders, attention mechanism, and multiple independent sets of learnable queries. However, the transformer decoder structure has not yet been well explored in the VPR task, despite it being highly renowned in the fields of NLP \cite{Transformer} and semantic segmentation \cite{tcsvtss1,senformer, maskformer, cheng2021mask2former} due to its strong capability in capturing contextual dependencies and producing accurate features. We argue that these characteristics of the decoder hold significant potential for addressing various challenges in VPR. To this end, we revisit the transformer decoder and propose an \textbf{E}fficient \textbf{D}ecoder \textbf{T}rans\textbf{former} (EDTformer), which can fully utilize the contextual information within deep features extracted from the backbone, then gradually decode and aggregate the crucial features to output robust and discriminative global representations for the place images.

Moreover, as vision foundation models \cite{dinov2,yuan2021florence,radford2021CLIP} have demonstrated a remarkable ability for feature extraction, recent VPR studies tend to use a pre-trained foundation model, e.g., DINOv2 \cite{dinov2}, as the backbone to extract deep features from the input place images. AnyLoc \cite{anyloc} first applied DINOv2 for VPR without any fine-tuning, making it difficult to fully unleash the capability of DINOv2 for VPR. Then some research (e.g., BoQ \cite{BoQ} and SALAD \cite{SALAD}) began to enhance the performance of DINOv2 in VPR through only fine-tuning the last few encoder blocks of DINOv2. Unfortunately, this process is accompanied by a large number of trainable parameters. Concurrently, CricaVPR \cite{cricavpr} and SelaVPR \cite{selavpr} attempted to apply parameter-efficient transfer learning (PETL) methods \cite{adapter,adapterformer,Convadapter} to adapt DINOv2 to the VPR task. Specifically, they froze DINOv2 and inserted additional trainable adapters into each encoder block. However, this approach is efficient only in terms of parameters, not in training time and memory usage, as the gradient computation for the trainable parameters still requires backpropagation through (nearly the entire) pre-trained backbone \cite{lst,2024LoSA,Diao2023UniPT}, as shown in Fig. \ref{fig:LoPA} (b). This motivates us to develop a method that can fully strengthen the performance of vision foundation models in the VPR task in a memory- and parameter-efficient way.

In this paper, we revisit the transformer decoder and propose a novel feature aggregation method EDTformer, which can finally produce robust and discriminative global representations for the place images. Without bells and whistles, to fully leverage the rich contextual information within deep features, EDTformer employs a cascade of our simplified decoder blocks. These blocks only retain the attention (self-attention and cross-attention) layers and remove the feedforward network for higher efficiency compared to the standard transformer decoder. By taking deep features as the keys and values, as well as introducing a set of learnable parameters as queries, EDTformer can capture complex contextual relationships encoded in the deep features and progressively decode effective features into the learnable queries to achieve principal feature aggregation. Subsequently, the learnable queries that have contained crucial information are processed through two simple linear layers for dimensionality reduction and further adjustment, producing the global representations. Meanwhile, we use the foundation model DINOv2 as the backbone of our framework and develop a memory- and parameter-efficient Low-rank Parallel Adaptation (LoPA) method to adapt DINOv2 for the VPR task. Specifically, we freeze the whole backbone during training and design a tunable lightweight parallel network, which can progressively refine the intermediate features output by each encoder block of DINOv2 to provide more powerful deep features for EDTformer, thus enhancing the robustness of our entire model.

The main contributions of our work are summarized as follows:

\textbf{(1)} We revisit the transformer decoder and propose a novel feature aggregation method EDTformer, by leveraging several stacked simplified decoder blocks, linear projection, and a set of learnable queries to fully decode the deep features and finally output a robust and discriminative global representation for global-retrieval-based VPR. This provides a new insight into how to apply the decoder structure for VPR.

\textbf{(2)} We design a Low-rank Parallel Adaptation method to adapt the foundation model DINOv2 to output enhanced features for boosting performance, which is efficient not only in terms of parameters, but also in training time and memory usage. This can further facilitate the application of the foundation models in resource-constrained VPR scenarios.

\textbf{(3)} Extensive experiments on the benchmark datasets show that our method can outperform the state-of-the-art (SOTA) VPR methods by a considerable margin with less memory usage. The results on multiple datasets which reflect the advantages of our method are shown in Fig. \ref{fig:R@1 Performance}.

\section{Related Work}
\subsection{Visual Place Recognition}
In early VPR research, global features were developed by aggregating the hand-crafted local features (e.g., SIFT \cite{SIFT} and SURF \cite{SURF1,SURF2}), employing some classical aggregation algorithms, such as Bag of Words \cite{BagofWords}, Fisher Vector \cite{fishvector} and Vector of Locally Aggregated Descriptors (VLAD) \cite{VLAD}. With the great success of deep learning in compute vision tasks, current predominant VPR methods \cite{netvlad,dlf,CRN,R-MAC,Tutorial,SFRS,patch-netvlad,transvpr,STA-VPR,garg2021place,tcsvtvpr1,benchmark,Cosplace,eigenplaces,anyloc,mixvpr,BoQ,SALAD,cricavpr,selavpr} have preferred leveraging large amounts of deep features rather than hand-crafted local features to boost performance. Besides, the traditional aggregation algorithms are gradually replaced by trainable aggregation/pooling layers, such as NetVLAD \cite{netvlad} and GeM pooling \cite{GeM}. NetVLAD utilizes a trainable generalized VLAD layer to aggregate deep features, typically tending to get a high-dimensional descriptor. In contrast, the Generalized Mean (GeM) pooling is a simple alternative that can produce compact global representations. However, such compact representations usually fall short of delivering satisfactory performance in challenging VPR scenarios. Hence, numerous works \cite{SFRS,densernet,Cosplace,eigenplaces,mixvpr,cricavpr,SALAD,BoQ} proposed various novel training strategies and aggregation algorithms to further improve the global representations for better retrieval accuracy. SFRS \cite{SFRS} proposed self-supervised image-to-region similarities to thoroughly exploit the potential of challenging positive images and their corresponding sub-regions for training a more robust VPR model. CosPlace \cite{Cosplace} and EigenPlaces \cite{eigenplaces} framed the task of VPR training as a classification problem and utilized the San Francisco eXtra Large (SF-XL) dataset to train their VPR models effectively. MixVPR \cite{mixvpr} proposed an all-MLP aggregation technique and trained the model utilizing the multi-similarity loss \cite{multi-similarity} with full supervision. CricaVPR \cite{cricavpr} introduced a cross-image correlation-aware representation learning method to enhance the robustness of global features. SALAD \cite{SALAD} reinterpreted the soft assignment of local features in NetVLAD as an optimal transport problem, solving it by using the Sinkhorn algorithm \cite{sinkhorn}. BoQ \cite{BoQ} utilized additional transformer encoders to further process deep features from the backbone, and then used multiple independent sets of queries to capture universal place-specific attributes from the output of each encoder through employing attention mechanism. These studies only used global features and achieved a relatively good performance.

In addition, two-stage VPR methods \cite{patch-netvlad,transvpr,zhu2023r2former,dhevpr}, which first search for the top-k candidate images in the database using global features and then re-ranks the candidates based on local features, are also an effective approach to further improve recognition performance. The re-ranking process usually either employs geometric consistency verification or leverages the learnable network to produce dense local features for similarity computation \cite{zhu2023r2former,dhevpr,selavpr}. For instance, DHE-VPR \cite{dhevpr} adopted a transformer-based deep homography estimation network to fit homography for fast and learnable geometric verification. SelaVPR \cite{selavpr} introduced a novel hybrid global-local adaptation method and directly used the dense local features in cross-matching for re-ranking. However, re-ranking is generally time-consuming and demands a huge memory footprint as well as substantial storage, particularly when dealing with large databases. These shortcomings restrict the applicability of two-stage methods in resource-constrained and large-scale VPR scenarios. Unlike two-stage methods improving performance at a substantial cost, our proposed method efficiently uses the decoder to aggregate deep features and directly produce highly robust and discriminative global representations, thus addressing the main challenges in VPR.

\subsection{Transformer Decoder Architecture}
Since the success of the transformer \cite{Transformer} in NLP, the scope of transformer decoder has expanded to various computer vision tasks \cite{tcsvtdecoder1,tcsvtdecoder2,detr,mask2former,maskformer}. For instance, DETR \cite{detr} was the first to apply the transformer decoder for object detection and used multiple learnable object queries to capture the information of target objects, finally outputting a set of predictions in parallel. SenFormer \cite{senformer}, MaskFormer \cite{maskformer} and Mask2Former \cite{mask2former} used the stacked transformer decoders to generate segmentation masks by learning query sets, which can refine multi-scales features extracted from the backbone and denote some possible segmentation regions. IAA-LQ \cite{xiong2024image} used a standard transformer decoder to estimate the aesthetics of images. Different from them, we first simplify the structure of the transformer decoder by removing the feedforward network to get a higher efficiency. Additionally, the purpose of using learnable queries is distinct. Our learnable queries primarily are used to aggregate global contextual information from deep features of the place image under the action of our EDTformer, rather than focusing on specific pixels or pixel blocks/objects in the image.

\subsection{Parameter-Efficient Transfer Learning}
The vision foundation models \cite{dinov2,radford2021CLIP,yuan2021florence} trained on huge quantities of data, such as DINOv2 (trained on the large-scale curated LVD-142M dataset with the self-supervised strategy), possess the ability to extract powerful features from the input images and have achieved remarkable results on various downstream vision tasks. In order to reduce the number of training parameters while maintaining the strong capability of these foundation models, the PETL methods \cite{adapter,adapterformer,lora,Convadapter} have been proposed and are widely applied in many areas, including the VPR field. For example, CricaVPR \cite{cricavpr} enhanced the performance of DINOv2 in VPR by introducing multi-scale convolution adapters. SelaVPR \cite{selavpr} inserted the vanilla adapters into DINOv2 to achieve a hybrid global-local adaptation to produce both global and local features for the VPR task. By only tuning the built-in lightweight adapters without adjustment to the frozen pre-trained model, they are both efficient in terms of parameter. Nevertheless, the memory overhead during training is also dominated by the activations, not only parameters, which means that parameter efficiency is not equivalent to memory efficiency \cite{lst,2024LoSA,Diao2023UniPT}. Unlike them, we design a Low-rank Parallel Adaptation method to fully unleash the capability of the pre-trained vision foundation model without inserting any parameter into it, which is both parameter- and memory-efficient. 

\begin{figure*}[t]
    \centering
    \vspace{-0.5cm}
    \includegraphics[width=\textwidth]{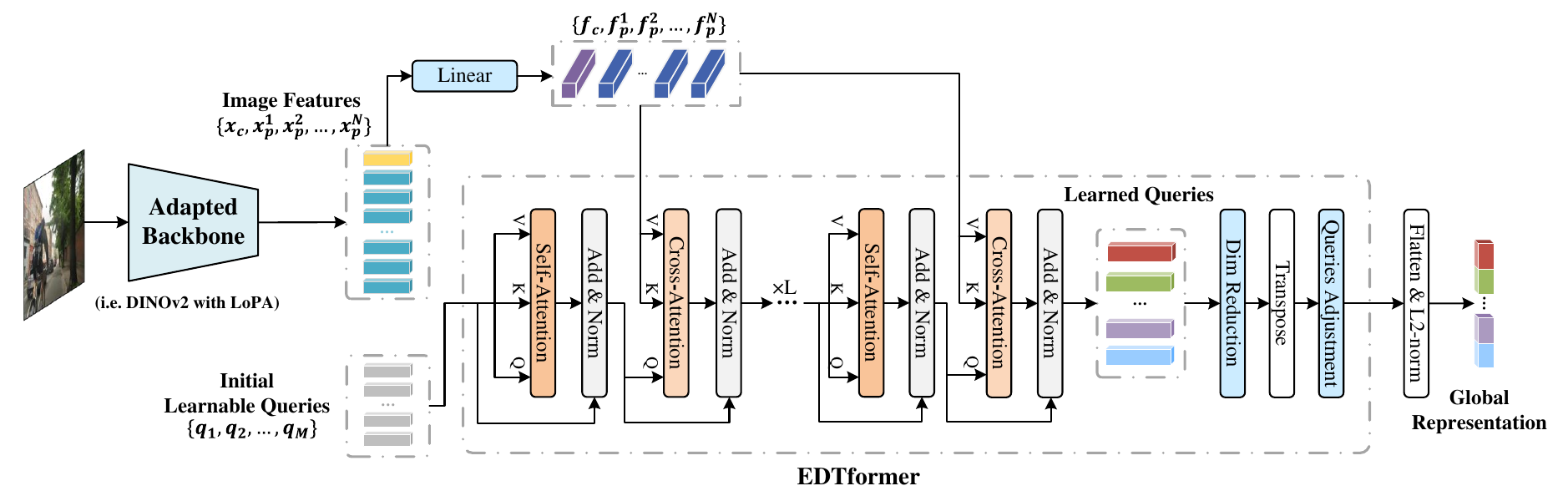}
    \caption{Our pipeline to produce the robust and discriminative global representation for single-stage VPR. Firstly, the frozen backbone with Low-rank Parallel Adaptation (i.e., DINOv2 with LoPA) is employed to extract powerful deep features of the input image. Next, the features undergo a linear transformation and are fed into each cross-attention layer as the keys and values. Additionally, we initial a set of learnable queries as the input queries for the first self-attention layer. After passing through $L$ our simplified decoder blocks, we can obtain the learned queries which have aggregated the crucial contextual features for the VPR task. Then we use two fully connected layers: one for dimensionality reduction and the other for further information aggregation by adjusting the number of queries. Finally, the output features are flattened and L2-normalized as the global representation of the place image.}
    \label{fig:pipeline}
\end{figure*}

\section{Methodology}
In this section, we first briefly introduce our backbone for feature extraction. Next, we propose the EDTformer method and describe in detail how it achieves feature aggregation efficiently. Then we introduce our LoPA method to adapt the foundation model in a memory- and parameter-efficient way to provide more powerful features for EDTformer, thus enhancing the robustness of our entire model. Finally, we describe the training strategy used in our experiments.

\subsection{Feature Extraction}
\begin{figure*}
    \centering
    \vspace{-0.5cm}
    \includegraphics[width=1.0\textwidth]{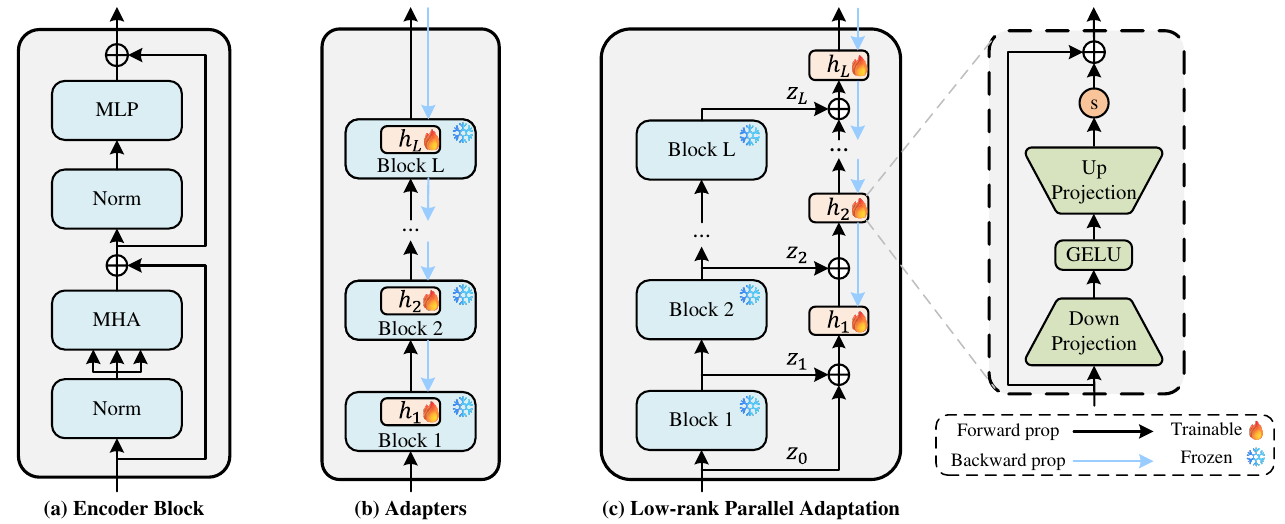}
    \caption{Illustration of our Low-rank Parallel Adaptation method. (a) is a standard transformer encoder block in ViT. (b) is the popular PETL method based on adapters, which usually inserts the trainable adapters into the encoder blocks of the frozen backbone. (c) is our proposed LoPA method. The intermediate features from each encoder block of the frozen DINOv2 are sequentially fed to the corresponding adaptation function together with the output from the previous adaptation function.} During training, backpropagation does not pass through the frozen backbone, thus greatly reducing memory usage.
    \label{fig:LoPA}
    \vspace{-0.2cm}
\end{figure*}

In this work, we adopt the vision foundation DINOv2 as the backbone, which is based on Vision Transformer (ViT) \cite{ViT}. To process the input image $I\in\mathbb{R}^{w \times h \times c}$, ViT first divides the image into $p \times p$ non-overlapping small patches, and then linearly projects them into $d$-dimensional tokens $x_p \in \mathbb{R}^{N \times d}$ ($N=p \times p$). Meanwhile, a learnable class token is prepended to $x_p$ to obtain $x_0=[x_{class},x_p] \in \mathbb{R}^{(N+1) \times d}$. Subsequently, position embeddings are added to $x_0$ to preserve positional information, resulting in $z_0$, which will be processed by a series of repeated transformer encoder blocks to extract features. A transformer encoder block mainly consists of the Multi-Head Attention (MHA) layer, the Multi-Layer Perceptron (MLP), and the LayerNormalization (LN), as shown in Fig. \ref{fig:LoPA} (a). The token sequence $z_{l-1}$ goes through the transformer encoder block and produces $z_l$, which can be formulated as
\begin{equation}
    z^{\prime}_l = MHA(LN(z_{l-1})) + z_{l-1},
\end{equation}
\begin{equation}
    z_l = MLP(LN(z^{\prime}_l)) + z^{\prime}_l,
\end{equation}
where $z_{l-1}$ and $z_l$ are the outputs of the $(l-1)$-th and $l$-th transformer encoder blocks, respectively. Based on these extracted features, we will further introduce how to adapt DINOv2 efficiently using LoPA in the subsection \ref{subsection: LoPA}.

\subsection{EDTformer}
The EDTformer primarily relies on MHA mechanism, so we provide a detailed introduction. The MHA first maps the input sequence into queries \textit{Q}, keys \textit{K}, and values \textit{V} \textit{h} times with different learnable linear projections. Then, the attention between \textit{Q}, \textit{K}, and \textit{V} is computed through scaled dot-product \cite{Transformer}, formulated as
\begin{equation}
    Attn(Q,K,V)=softmax\big(QK^\top/\sqrt{d}\big)V.
\end{equation}
On each of these projected versions of queries, keys and values, we then perform the attention function in parallel, yielding \textit{h} output values. These are concatenated and once again projected ($W_O$), resulting in the final values, which can be expressed as
\begin{equation}
    MHA(Q,K,V) = Concat(head_1,\dots,head_h)W_O,
\end{equation}
\begin{equation}
    head_i = Attn(Q_i,K_i,V_i).
\end{equation}
Notably, the self-attention layer uses the same feature as query, key and value, whereas the cross-attention layer uses one feature as query and the other feature as key and value. We will use both of them to construct EDTformer to obtain robust and discriminative global features.

We design a simple yet powerful pipeline to produce promising global features as shown in Fig. \ref{fig:pipeline}. For the deep features $\mathcal{X}: \{x_c,x^{1}_{p}, x^{2}_{p}, \ldots, x^{N}_{p}\}\in \mathbb{R}^{(N+1) \times d}$ output by the adapted backbone (i.e., DINOv2 with LoPA), we apply a linear projection to $\mathcal{X}$ for feature transformation and information transfer to obtain features $\mathcal{F}: \{f_c, f^{1}_{p}, f^{2}_{p}, \ldots, f^{N}_{p}\} \in \mathbb{R}^{(N+1) \times d}$, which can be formulated as 
\begin{equation}
    \mathcal{F} = W_1\mathcal{X} + b_1.
\end{equation}
The proposed EDTformer adopts a pure decoder structure, which utilizes the simplified decoder block and a fixed set of learnable queries. Different from the standard transformer decoder, our simplified decoder only consists of a self-attention layer and a cross-attention layer without the feedforward network (FFN) for higher efficiency. 
For the first self-attention layer, the input vector \textit{Q}, \textit{K}, \textit{V} come from a set of learnable queries, denoted as $\mathcal{Q}: \{q_1, q_2, \dots, q_M\} \in \mathbb{R}^{M \times d}$ ($M < N$). These queries are trainable parameters of the model and independent of the input features. Notably, for the subsequent self-attention layers, the input comes from the output of the preceding decoder block. Through the self-attention operation, these queries can conduct internal information interaction to reduce redundancy and highlight the essential parts. 
Unlike the self-attention layer, the input of the cross-attention layer comes from two sources: \textit{Q} is derived from the output of the self-attention layer, while \textit{K} and \textit{V} are the previous features $\mathcal{F}$. By the cross-attention mechanism, the module can dynamically transfer and aggregate effective contextual information within $\mathcal{F}$ to the learnable queries $\mathcal{Q}$ according to the cross-attention weights between them. To fully make use of $\mathcal{F}$, we adopt $L$ decoder blocks in a cascading way to decode layer by layer, thereby the initial learnable queries $\mathcal{Q}$ gradually result in the learned queries $\mathcal{O}_L: \{o_1, o_2, \dots, o_M\}\in \mathbb{R}^{M \times d}$. The internal process of each decoder block can be denoted as
\begin{equation}
    \mathcal{Q}_i = LN(MHA(\mathcal{O}_{i-1},\mathcal{O}_{i-1},\mathcal{O}_{i-1})+\mathcal{O}_{i-1}),
\end{equation}
\begin{equation}
    \mathcal{O}_i = LN(MHA(\mathcal{Q}_i,\mathcal{F},\mathcal{F})+\mathcal{Q}_i),
\end{equation}
where $\mathcal{Q}_i$ and $\mathcal{O}_i$ respectively are the outputs of the $i$-th self-attention layer and cross-attention layer. Notably, the $\mathcal{O}_0$ is the initial learnable queries $\mathcal{Q}$.

Subsequently, we utilize two fully connected layers: the first to reduce the dimensionality of learned queries, and the second to adjust the number of queries for further feature aggregation, formulated as 
\begin{equation}
    \text{Output} = W_3(W_2\mathcal{O}_L + b_2)^T + b_3.
\end{equation}
Finally, we flatten the output features and employ the L2-normalization to obtain the robust and discriminative global representation of the input place image.

It is worth noting that the latest work BoQ \cite{BoQ} and our EDTformer both utilize learnable queries during feature aggregation, but our architectures are different. BoQ uses several additional transformer encoders to further deal with the deep features extracted from the backbone and applies separate attention mechanism, as well as multiple independent sets of queries to individually learn the output of each encoder. In contrast, we first attempt to directly utilize a purely decoder-based structure with only a set of learnable queries to consistently decode and aggregate the effective information from deep features, thus obtaining a robust and discriminative global representation, which is simpler and more efficient.

\subsection{Low-rank Parallel Adaptation}
\label{subsection: LoPA}
The foundation model DINOv2 recently has become a popular backbone for feature extraction in VPR, generally accompanied by a fine-tuning/adaptation for better performance. CricaVPR \cite{cricavpr} and SelaVPR \cite{selavpr} have used the PETL methods (i.e., inserting tunable adapters into the frozen encoder blocks) to adapt DINOv2, as shown in Fig. \ref{fig:LoPA} (b). However, they are efficient only in terms of parameters, not in training time and memory usage. Here, we conduct a simple theoretical derivation to explain this point. For a network with $L$ blocks, each having output ${z}_i$, a large number of frozen parameters $\theta_i$ and a few trainable parameters $\psi_i$, our goal is to minimize a loss function $\mathcal{L}$ based on stochastic gradient descent. Concretely, to update $\psi_i$ during backpropagation, we first need to compute $\frac{\partial \mathcal{L}}{\partial \psi_i}=\frac{\partial \mathcal{L}}{\partial z_i} \frac{\partial z_i}{\partial \psi_i}$. However, $\frac{\partial \mathcal{L}}{\partial z_i}$ cannot be computed directly, because $\mathcal{L}$ is usually calculated
using the final output of the network (i.e, $z_L$). Therefore, it is necessary to resort to the chain rule as follows, which is the core of the famous backpropagation algorithm
\begin{equation}
\frac{\partial \mathcal{L}}{\partial \psi_i} = \frac{\partial \mathcal{L}}{\partial z_i} \frac{\partial z_i}{\partial \psi_i} = \underbrace{\frac{\partial \mathcal{L}}{\partial z_L} \frac{\partial z_L}{\partial z_{L-1}} \cdots \frac{\partial z_{i+1}}{\partial z_i}}_{\text{intermediate activations}} \frac{\partial z_i}{\partial \psi_i}.
\label{eq: chain}
\end{equation}
Although most parameters (i.e., $\theta_i$) in each block are frozen, it is inevitable to compute gradients about the intermediate activations to assist in updating these few trainable parameters (i.e., $\psi_i$), as shown in Eq. (\ref{eq: chain}), which incurs huge memory overhead and additional running time.

To tackle the above issue, we propose the Low-rank Parallel Adaptation method inspired by previous studies \cite{lst,2024LoSA,Diao2023UniPT}. Instead of inserting tunable parameters inside the backbone, we design a parallel network that does not require backpropagation gradients through the frozen backbone during training as shown in Fig. \ref{fig:LoPA} (c). It is a lightweight and separate network, which directly takes the intermediate features produced by each encoder block of the backbone as inputs and refines them progressively to output more accurate representations. 

Concretely, our backbone DINOv2 consists of a patch embedding block and $L$ encoder blocks, and therefore $L+1$ outputs $z_0,z_1,z_2...,z_L$, each composed of $N+1$ tokens with a dimensionality of $d$, so $z_i \in \mathbb{R}^{(N+1) \times d}$. We learn parallel adaptation functions, $h$, which operate on these intermediate features to refine them. We denote the outputs of our adaptation functions as $y_i$ where $i$ denotes the function index. The whole process can be formulated as 
\begin{equation}
    y_i = \begin{cases}
        h_i(z_{i-1} + z_i) & \text{if } i = 1, \\
        h_i(y_{i-1} + z_i) & \text{otherwise}.
    \end{cases}
\end{equation}
To be more efficient, we design our adaptation function $h$ in a low-rank structure with few parameters. Specially, $h$ consists of a down-projection, $W_d: \mathbb{R}^d \rightarrow \mathbb{R}^r $, a GeLU non-linearity activation $\sigma(\cdot)$ \cite{hendrycks2016gelus}, and an up-projection, $W_u: \mathbb{R}^r \rightarrow \mathbb{R}^d$ with a skip-connection, where $r \ll d$. As in \cite{selavpr,cricavpr,2024LoSA}, we also add a scaling factor $s$, meaning that our adaptation function can be denoted as 
\begin{equation}
    h(x) = s W_u\sigma(W_dx) + x.
    \label{parallel_adapter}
\end{equation}
Notably, our LoPA adapts the foundation model for the VPR task in a parameter- and memory-efficient way, thus providing more powerful features for subsequent aggregator and enhancing the robustness of the entire model. In other words, it is also applicable to other VPR methods when using the foundation model, instead of being tailored only for our EDTformer.

\subsection{Training Strategy} 
The proposed model is trained on the GSV-Cities \cite{gsv_cities} dataset following its standard framework. GSV-Cities contains 560k images with highly accurate labels depicting 67k different places. Specifically, each place includes a set of images that describe the same location with different viewpoints and conditions. Besides, a unique label is assigned to each place. That is, the place images in GSV-Cities are divided into a restricted number of categories. Based on this characteristic, we apply the multi-similarity (MS) loss \cite{multi-similarity} function with an online hard mining strategy. The MS loss is formulated as 
\begin{equation}
\begin{aligned}\mathcal{L}_{MS} &= \frac{1}{B} \sum_{q=1}^{B} \left\{ \frac{1}{\alpha} \log \left[ 1 + \sum_{p \in \mathcal{P}_q} e^{-\alpha (S_{qp} - \lambda)} \right] \right. \\
                &\quad\left. + \frac{1}{\beta} \log \left[ 1 + \sum_{n \in \mathcal{N}_q} e^{\beta (S_{qn} - \lambda)} \right] \right\},
\end{aligned}
\label{MS_LOSS}
\end{equation}
where for each query image $I_q$ in a batch, $\mathcal{P}_q$ represents the set of indices \{$p$\} which align with the positive samples for $I_q$, and $\mathcal{N}_q$ represents the set of indices \{$n$\} which align with the negative samples for $I_q$. $S_{qp}$ and $S_{qn}$ respectively represent the cosine similarity of a positive sample pair $\{I_q, I_p\}$ and a negative sample pair $\{I_q, I_n\}$. The remaining variables $\alpha$, $\beta$ and $\lambda$ are three sets of hyperparameters.

\section{Experiments}
\subsection{Datasets and Evaluation Metrics}
To demonstrate the effectiveness of our method, we conduct the experiments on multiple benchmark datasets, which present a diverse set of challenges encountered in the real world. Table \ref{tab:description of datasets} provides a concise summary of these datasets. \textbf{Pitts30k} \cite{pitts30k}, extracted from Google Street View, shows significant changes in viewpoint. \textbf{MSLS} \cite{msls} is collected from 30 major cities across six continents over nine years period, covering all seasons and encompassing diverse weather, cameras, daylight conditions, and structural settings. \textbf{Tokyo24/7} \cite{tokyo247} presents severe lighting (day/night) variations. \textbf{SPED} \cite{sped} is collected from surveillance cameras and consists of various low-quality images. \textbf{Nordland} \cite{nordland} is captured using a front-mounted train camera, spanning all four seasons. \textbf{AmsterTime} \cite{amstertime} exhibits substantial image modality variations, using historical grayscale images as queries and contemporary color images as database. \textbf{SVOX} \cite{svox} is a cross-domain VPR dataset gathered under diverse weather and lighting conditions. We primarily utilize the three most challenging query subsets: SVOX Night, SVOX Rain and SVOX Sun. All these datasets are organized following the widely-used visual geo-localization benchmark \cite{benchmark}.

We use the Recall@N (R@N) in the experiments for performance evaluation, which is defined as the percentage of query images for which at least one of the top-N candidates falls within a threshold of ground truth. Consistent with previous work \cite{pitts30k,msls,tokyo247,cricavpr,eigenplaces}, we set the threshold to 25 meters and $40^\circ \text{angle}$ for MSLS, 25 meters for Pitts30k, Tokyo24/7, SPED and SVOX, $\pm 10$ frames for Nordland, special counterpart for AmsterTime.
\begin{table}
    \centering
    \caption{Brief Summary of the Benchmark Datasets in Our Experiments.}
    \label{tab:description of datasets}
    \resizebox{\columnwidth}{!}{%
    \begin{tabular}{ l c c c }
        \toprule
        \multirow{2}{*}{Dataset} & \multirow{2}{*}{Description} & \multicolumn{2}{c}{Number} \\
        \cmidrule{3-4}
        & & Database & Queries \\
        \midrule
        Pitts30k \cite{pitts30k} & viewpoint changes & 10,000 & 6,816 \\
        MSLS-val \cite{msls} & urban, suburban & 18,871 & 740 \\
        MSLS-challenge \cite{msls} & long-term & 38,770 & 27,092 \\
        Tokyo24/7 \cite{tokyo247} & illumination changes & 75,984 & 315 \\
        SPED \cite{sped} & various scenes & 607 & 607 \\
        Nordland \cite{nordland} & season variants & 27,592 & 27,592 \\
        Amstertime \cite{amstertime} & domain variants & 1,231 & 1,231 \\
        SVOX \cite{svox} & condition variations & 17,166 & 14,278 \\
        \bottomrule
    \end{tabular}%
    }
\end{table}
\subsection{Implementation Details}
We use DINOv2-base as the backbone and conduct experiments on NVIDIA GeForce RTX 4090 GPUs using PyTorch. The backbone is completely frozen and only LoPA is trainable to refine the features from the backbone progressively. The low rank $r$ in LoPA is set to 4 and the scaling factor $s$ in Eq. (\ref{parallel_adapter}) is set to 0.5. We apply two stacked simplified decoder blocks for a trade-off between accuracy and efficiency, and leverage 64 queries to fully learn the effective features. Finally, the model outputs a 4096-dim global descriptor. We set the hyperparameters $\alpha = 1$, $\beta = 50$, $\lambda = 0$ in Eq. (\ref{MS_LOSS}) and margin = 0.1 in online mining, as in previous work \cite{mixvpr,cricavpr,BoQ}. We train our model using the Adam optimizer with the initial learning rate set as 0.0001 and multiplied by 0.7 after every 3 epochs. Each training batch consists of 72 places, with 4 images per place, totaling 288 images. The resolution of the input image is $224\times224$ during training ($322\times322$ in reference). We train the model for 15 epochs in total.
\begin{table*}[ht]
    \centering
    \vspace{-0.5cm}
    \caption{Comparison to SOTA Methods on Benchmark Datasets. The Best Is Highlighted in Bold and the Second Is Underlined. $\dagger$ CricaVPR and SuperVLAD Both Utilize the Cross-Image Encoder to Correlate Multiple Images from the Same Place at Once in Inference to Get a Better Performance. They Are not Included in the Comparison with Other Methods. We Additionally Show the Results (CricaVPR-1 and SuperVLAD-1) of Single Query Image in Inference. $\ddag$ We Reproduce the Results of BoQ by Strictly Following Its Training Pipeline, Except for Keeping the Same Image Size for Both Training ($224\times224$) and Inference ($322\times322$) as Our Method.}
    \label{tab:comparison with state-of-the-art}
    \setlength{\tabcolsep}{0.5mm}
    \begin{tabularx}{\textwidth}{l|c|XXX|XXX|XXX|XXX|XXX}
        \toprule
        \multirow{2}{*}{Method} & \multirow{2}{*}{Dim} & \multicolumn{3}{c|}{MSLS-challenge} & \multicolumn{3}{c|}{Tokyo24/7} & \multicolumn{3}{c|}{MSLS-val} & \multicolumn{3}{c|}{SPED} & \multicolumn{3}{c}{Pitts30k} \\ 
        \cmidrule{3-5} \cmidrule{6-8} \cmidrule{9-11} \cmidrule{12-14} \cmidrule{15-17}
        & & R@1 & R@5 & R@10 & R@1 & R@5 & R@10 & R@1 & R@5 & R@10 & R@1 & R@5 & R@10 & R@1 & R@5 & R@10 \\
        \midrule
         NetVLAD \cite{netvlad} & 32768 & 35.1 & 47.4 & 51.7 & 60.6 & 68.9 & 74.6 & 53.1 & 66.5 & 71.1 & 70.2 & 84.5 & 89.5 & 81.9 & 91.2 & 93.7 \\
         SRFS \cite{SFRS} & 4096 & 41.6 & 52.0 & 56.3 & 81.0 & 88.3 & 92.4 & 69.2 & 80.3 & 83.1 & 80.2 & 92.6 & 95.4 & 89.4 & 94.7 & 95.9 \\
         CosPlace \cite{Cosplace} & 512 & 61.4 & 72.0 & 76.6 & 81.9 & 90.2 & 92.7 & 82.8 & 89.7 & 92.0 & 75.5 & 87.0 & 89.6 & 88.4 & 94.5 & 95.7 \\
         MixVPR \cite{mixvpr} & 4096 & 64.0 & 75.9 & 80.6 & 85.1 & 91.7 & 94.3 & 88.0 & 92.7 & 94.6 & 85.2 & 92.1 & 94.6 & 91.5 & 95.5 & 96.3 \\
         EigenPlaces \cite{eigenplaces} & 2048 & 67.4 & 77.1 & 81.7 & 93.0 & 96.2 & 97.5 & 89.1 & 93.8 & 95.0 & 82.4 & 91.4 & 94.7 & 92.5 & \underline{96.8} & 97.6 \\
         AnyLoc \cite{anyloc} & 49152 & 42.2 & 53.5 & 58.1 & 87.3 & 96.8 & 97.5 & 68.7 & 78.2 & 81.8 & 85.3 & 94.4 & 95.4 & 87.0 & 94.3 & 96.7 \\
         CricaVPR$^\dagger$ \cite{cricavpr} & 4096 & 69.0 & 82.1 & 85.7 & 93.0 & 97.5 & 98.1 & 90.0 & 95.4 & 96.4 & 91.4 & 95.6 & 96.7 & 94.9$^\dagger$ & 97.3$^\dagger$ & 98.2$^\dagger$ \\
         CricaVPR-1 \cite{cricavpr} & 4096 & 66.9 & 79.3 & 82.3 & 89.5 & 94.6 & 96.2 & 88.5 & 95.1 & 95.7 & 87.3 & 92.9 & 94.7 & 91.6 & 95.7 & 96.9 \\
         SuperVLAD$^\dagger$ \cite{supervlad} & 3072 & 75.3 & 86.8 & 89.9 & 95.2 & 97.8 & 98.1 & 92.2$^\dagger$ & 96.6$^\dagger$ & 97.4$^\dagger$ & 93.2$^\dagger$ & 97.0$^\dagger$ & 98.0$^\dagger$ & 95.0$^\dagger$ & 97.4$^\dagger$ & 98.2$^\dagger$ \\
         SuperVLAD-1 \cite{supervlad} & 3072 & 74.2 & 85.4 & 88.8 & 94.3 & \underline{97.8} & \underline{98.1} & 91.2 & 95.9 & 96.9 & 90.0 & 95.1 & 95.9 & 92.3 & 96.7 & \underline{97.8} \\
         BoQ$^\ddag$ \cite{BoQ} & 12288 & \underline{75.9} & 87.4 & 90.3 & \underline{95.2} & \underline{97.8} & \underline{98.1} & 91.2 & 95.7 & 96.4 & 88.6 & 95.2 & 96.2 & 91.9 & 95.9 & 97.2 \\
         SALAD \cite{SALAD} & 8448 & 75.0 & \underline{88.8} & \underline{91.3} & 94.6 & 97.5 & 97.8 & \textbf{92.2} & \underline{96.4} & \underline{97.0} & \underline{92.1} & \textbf{96.2} & \underline{96.6} & 92.5 & 96.4 & 97.5 \\
         SelaVPR (global) \cite{selavpr} & 1024 & 69.6 & 86.9 & 90.1 & 81.9 & 94.9 & 96.5 & 87.7 & 95.8 & 96.6 & 83.9 & 91.3 & 93.6 & 90.2 & 96.1 & 97.1 \\
         \midrule
         Patch-NetVLAD \cite{patch-netvlad} & / & 48.1 & 57.6 & 60.5 & 86.0 & 88.6 & 90.5 & 79.5 & 86.2 & 87.7 & 87.2 & 93.1 & 94.2 & 88.7 & 94.5 & 95.9 \\
         $R^2$Former \cite{zhu2023r2former} & / & 73.0 & 85.9 & 88.8 & 88.6 & 91.4 & 91.7 & 89.7 & 95.0 & 96.2 & 67.6 & 75.8 & 78.4 & 91.1 & 95.2 & 96.3 \\ 
         SelaVPR \cite{selavpr} & / & 73.5 & 87.5 & 90.6 & 94.0 & 96.8 & 97.5 & 90.8 & \underline{96.4} & \textbf{97.2} & 89.0 & 94.6 & 96.4 & \underline{92.8} & \underline{96.8} & 97.7 \\
         \midrule
         EDTformer (ours) & 4096 & \textbf{78.4} & \textbf{89.8} & \textbf{91.9} & \textbf{97.1} & \textbf{98.1} & \textbf{98.4} & \underline{92.0} & \textbf{96.6} & \textbf{97.2} & \textbf{92.4} & \underline{95.9} & \textbf{96.9} & \textbf{93.4} & \textbf{97.0} & \textbf{97.9} \\
         \bottomrule
    \end{tabularx}
\end{table*}

\begin{table*}[htbp]
    \centering
    \vspace{-0.3cm}
    \caption{Comparison to SOTA Methods on More Challenging Datasets. The Best Is Highlighted in Bold and the Second Is Underlined.}
    \label{tab:Performance in extreme scenarios}
    \setlength{\tabcolsep}{0.5mm}
    \begin{tabularx}{\textwidth}{l|XXX|XXX|XXX|XXX|XXX}
    \toprule
    \multirow{2}{*}{Method} & \multicolumn{3}{c|}{Nordland} & \multicolumn{3}{c|}{AmsterTime} & \multicolumn{3}{c|}{SVOX Night} & \multicolumn{3}{c|}{SVOX Rain} & \multicolumn{3}{c}{SVOX Sun}  \\
    \cmidrule{2-4} \cmidrule{5-7} \cmidrule{8-10} \cmidrule{11-13} \cmidrule{14-16}
    & R@1 & R@5 & R@10 & R@1 & R@5 & R@10 & R@1 & R@5 & R@10 & R@1 & R@5 & R@10 & R@1 & R@5 & R@10 \\
    \midrule
    SFRS \cite{SFRS} & 16.0 & 24.1 & 28.7 & 29.7 & 48.5 & 55.6 & 28.6 & 40.6 & 46.4 & 69.7 & 81.5 & 84.6 & 54.8 & 68.3 & 74.1 \\
    CosPlace \cite{Cosplace} & 58.5 & 73.7 & 79.4 & 38.7 & 61.3 & 67.3 & 44.8 & 63.5 & 70.0 & 85.2 & 91.7 & 93.8 & 67.3 & 79.2 & 83.8 \\
    MixVPR \cite{mixvpr} & 76.2 & 86.9 & 90.3 & 40.2 & 59.1 & 64.6 & 64.4 & 79.2 & 83.1 & 91.5 & 97.2 & 98.1 & 84.8 & 93.2 & 94.7 \\
    EigenPlaces \cite{eigenplaces} & 71.2 & 83.8 & 88.1 & 48.9 & 69.5 & 76.0 & 58.9 & 76.9 & 82.6 & 90.0 & 96.4 & 98.0 & 86.4 & 95.0 & 96.4 \\
    AnyLoc \cite{anyloc} & 26.3 & 41.0 & 48.8 & 38.6 & 61.7 & 69.8 & 65.2 & 82.5 & 89.8 & 79.5 & 93.0 & 95.9 & 79.4 & 93.6 & 95.9 \\
    CricaVPR-1 \cite{cricavpr} & 79.4 & 90.1 & 93.3 & 49.4 & 70.3 & 76.7 & 76.8 & 88.0 & 92.3 & 93.5 & 98.5 & 99.0 & 87.8 & 97.2 & 97.9 \\
    SuperVLAD-1 \cite{supervlad} & 85.6 & \underline{94.0} & \underline{96.3} & \underline{54.3} & \underline{76.8} & \underline{82.3} & 91.9 & 97.8 & 98.4 & 96.7 & 98.8 & 99.4 & 97.3 & 99.2 & 99.3 \\
    BoQ$^\ddag$ \cite{BoQ} & 85.0 & 93.5 & 95.7 & 52.8 & 74.2 & 80.2 & \underline{95.8} & \textbf{99.2} & \textbf{99.3} & \underline{97.8} & \underline{99.5} & \underline{99.7} & \underline{97.8} & \underline{99.4} & \underline{99.4} \\
    SelaVPR \cite{SALAD} & \underline{87.3} & 93.8 & 95.6 & 53.6 & 72.8 & 78.1 & 88.9 & 95.7 & 97.3 & 93.8 & 98.4 & 98.9 & 90.9 & 96.0 & 96.8 \\
    \midrule
    EDTformer (Ours) & \textbf{88.3} & \textbf{95.3} & \textbf{97.0} & \textbf{65.2} & \textbf{85.0} & \textbf{89.0} & \textbf{96.2} & \underline{98.7} & \textbf{99.3} & \textbf{98.7} & \textbf{99.8} & \textbf{99.8} & \textbf{98.5} & \textbf{99.5} & \textbf{99.8} \\
    \bottomrule
    \end{tabularx}
    \vspace{-0.4cm}
\end{table*}
\subsection{Comparison with State-of-the-Art Methods}
In this section, we compare our proposed method with a wide range of existing SOTA VPR algorithms, including ten single-stage methods using global features for direct retrieval: NetVLAD \cite{netvlad}, SFRS \cite{SFRS}, CosPlace \cite{Cosplace}, MixVPR \cite{mixvpr}, EigenPlaces \cite{eigenplaces}, AnyLoc \cite{anyloc}, CricaVPR \cite{cricavpr}, SuperVLAD \cite{supervlad}, BoQ \cite{BoQ} and SALAD \cite{SALAD}, as well as three superior two-stage methods: Patch-NetVLAD \cite{patch-netvlad}, $R^2$Former \cite{zhu2023r2former} and SelaVPR \cite{selavpr}. Notably, our work uses the same training dataset as SALAD and BoQ, i.e., GSV-Cities. Additionally, the latest works CricaVPR, SuperVLAD, SALAD, BoQ and SelaVPR all leverage the foundation model DINOv2 as the backbone (SelaVPR using DINOv2-large, while others using DINOv2-base) and achieve the SOTA performance on multiple benchmarks. Thus, we also apply the DINOv2-base. Table 
\ref{tab:comparison with state-of-the-art} shows the quantitative results on MSLS, Tokyo24/7, SPED and Pitts30k. Our method achieves the best R@1/R@5/R@10 performance on almost all datasets.

\textbf{Results analysis.} SALAD, BoQ, SelaVPR and our method all achieve superior performance on these datasets. Especially on Tokyo24/7, which shows severe illumination changes, SALAD, BoQ and SelaVPR achieve 94.6\% R@1, 95.2\% R@1 and 94.0\% R@1 respectively. However, our method consistently improves performance on Tokyo24/7, achieving an incredible 97.1\% R@1. This improvement stems from the robust and discriminative global descriptors produced by our model. The MSLS dataset is highly challenging, covering different seasons, weather and illumination conditions, various camera types and viewpoints, as well as different levels of dynamic objects presented in the scenes. Nevertheless, our method achieves 78.4\% R@1, 89.8\% R@5 and 91.9\% R@10 on MSLS-challenge, showing significant advantages and outperforming other global-retrieval-base methods and two-stage methods by a considerable margin. Although our method only achieves 92.0\% R@1 on MSLS-val, 0.2\% lower than SALAD, the dimensionality of our global descriptors is less than half that of SALAD. This indicates that our global descriptors require less storage space and can achieve higher retrieval efficiency. In addition, we also get an overall better performance on the SPED dataset, demonstrating the high robustness of our method to process low-quality place images. On the Pitts30k dataset, which shows severe viewpoint changes, our EDTformer achieves a competitive performance, surpassing BoQ and SALAD by 1.5\% and 0.9\% on R@1 respectively. Although our EDTformer is slightly behind CricaVPR and SuperVLAD on individual datasets, this is because CricaVPR and SuperVLAD both use the cross-image encoder to combine multiple query images from the same place in a batch to achieve better performance. As the number of query images in the batch (i.e., batch size) decreases, especially when it becomes 1, their performance drops significantly, as shown in Table \ref{tab:comparison with state-of-the-art} CricaVPR-1 and SuperVLAD-1.

To further evaluate the generalization of our method in some extreme scenarios, we conduct extensive experiments on other challenging datasets: Nordland, AmsterTime, SVOX Night, SVOX Rain and SVOX Sun. The results shown in Table \ref{tab:Performance in extreme scenarios} demonstrate the powerful capability of our method to effectively tackle the VPR task in difficult scenes. On the Nordland dataset, which exhibits significant variations in seasons and illumination, our EDTformer gets the best performance, even outperforming the two-stage SelaVPR. Additionally, our method surpasses other methods on the AmsterTime dataset by a large margin. Specifically, it outperforms the second by 10.9\%, 8.2\% and 6.7\% on R@1/R@5/R@10 respectively. This implies the high robustness of our method to handle image modality variations in the datasets both containing grayscale and colorful images. On the SVOX Night, SVOX Rain and SVOX Sun datasets, our EDTformer achieves a relatively perfect performance (e.g., 98.7\% R@1, 99.8\% R@5 and 99.8\% R@10 on SVOX Rain), which shows that it can be well applied to these special scenarios. In summary, the results on these challenging datasets further highlight the robust generalization capability of our model.

\begin{figure}[t]
    \centering
    \includegraphics[width=0.95\linewidth]{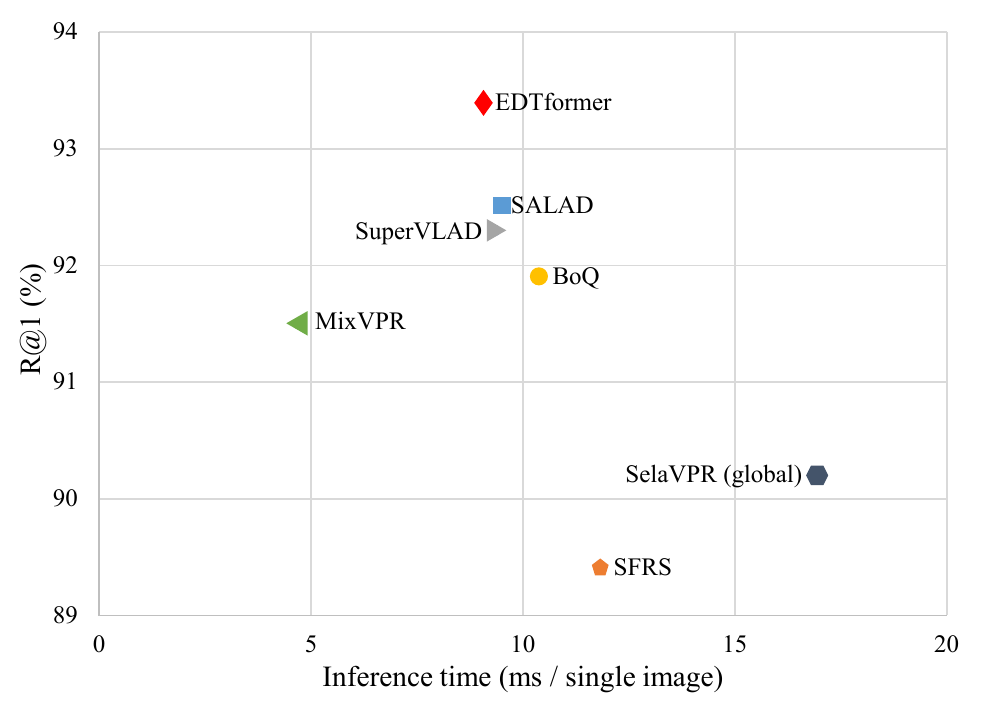}
    \caption{The R@1 and inference time comparison of different single-stage methods on Pitts30k. We consistently measure the inference time on an NVIDIA GeForce RTX 4090 GPU.}
    \label{fig:R@1_InferenceTime}
\end{figure}

\begin{table}[t]
    \centering
    \setlength{\tabcolsep}{0.8mm} 
    \caption{The Comparison of SALAD, BoQ, SuperVLAD and EDTformer in Training Memory Usage and Training Time. We Train the Four Models on GSV-Cities with Batch Size Set to 36 and Measure the Metrics on an NVIDIA GeForce RTX 4090 GPU.}
    \label{tab:Time_Memory}
    \begin{tabular}{c c c c c}
        \toprule
        Metric & SALAD & BoQ & SuperVLAD & EDTformer \\
        \midrule
        Training Memory Usage (GB) & 14.81 & 15.21 & 10.23 & \textbf{5.72} \\
        Training Time (min / epoch) & 11.21 & 10.67 & 10.57 & \textbf{8.77} \\
        \bottomrule
    \end{tabular}
\end{table}

Moreover, our method also exhibits an advantage in terms of efficiency in multiple aspects. Fig. \ref{fig:R@1_InferenceTime} exhibits the R@1 performance on the Pitts30k dataset and the inference time of a single image about seven global-retrieval-based methods, including SFRS, MixVPR, SALAD, BoQ, SuperVLAD, SelaVPR (global) and our EDTformer. MixVPR uses the CNN model (ResNet-50) as the backbone and proposes a feature mixing method to get global descriptors, which achieves the fastest inference speed. Although SFRS also leverages the CNN model (VGG-16), it applies PCA to reduce the dimensionality of the global descriptors, which is time-consuming. The other methods are all based on the vision foundation models. Specifically, SelaVPR (global) adopts DINOv2-large as the backbone, resulting in the slowest inference speed (16.95 ms). Among SALAD, BoQ, SuperVLAD and our EDTformer, all of which use DINOv2-base as the backbone, our method achieves the fastest inference speed (9.16 ms). In addition, we also provide the training memory consumption and training time per epoch for these four methods, as shown in Table. \ref{tab:Time_Memory}. Our training speed is approximately 15\% faster. Besides, our training memory usage is only about half that of SuperVLAD and one-third that of SALAD and BoQ. In summary, our method is efficient both in training and inference while achieving a superior recognition performance.

\subsection{Ablation Studies}
\label{sec: ablation studies}
In this section, we first conduct a series of ablation experiments to demonstrate the effectiveness of our proposed EDTformer and LoPA. Subsequently, we further explore the impact of some design details. Notably, we consistently use GSV-Cities with the MS loss for training in the ablation experiments. Unless otherwise specified, the dimensionality of the global features produced by our method is 4096. 

\begin{table}[t]
    \centering
    \caption{Ablation on Different Aggregation Methods. The Dimensionality of Global Descriptors Is Reported.}
    \label{tab:different aggregation methods}
    \setlength{\tabcolsep}{0.6mm}
    \begin{tabular}{l|c|c c | c c | c c | c c}
    \toprule
    \multirow{2}{*}{Method} &\multirow{2}{*}{Dim} & \multicolumn{2}{c|}{MSLS-val} & \multicolumn{2}{c|}{SPED} & \multicolumn{2}{c|}{Pitts30k} & \multicolumn{2}{c}{Nordland} \\
    \cmidrule{3-4} \cmidrule{5-6} \cmidrule{7-8} \cmidrule{9-10}
    & & R@1 & R@5 & R@1 & R@5 & R@1 & R@5 & R@1 & R@5 \\
    \midrule
    GeM & 768 & 85.7 & 94.6 & 76.1 & 88.5 & 88.5 & 94.4 & 32.3 & 47.1 \\
    GeM-linear & 4096 & 87.3 & 93.6 & 86.0 & 93.6 & 91.5 & 95.9 & 63.0 & 77.3 \\
    NetVLAD & 24576 & 89.7 & 95.5 & 88.8 & 95.1 & 92.0 & 96.5 & 71.7 & 84.8 \\
    NetVLAD (PCA) & 4096 & 89.3 & 95.3 & 87.5 & 93.9 & 92.0 & 96.3 & 71.7 & 84.6 \\
    Conv-AP & 4096 & 72.7 & 80.9 & 82.0 & 92.1 & 89.5 & 94.8 & 50.8 & 67.6 \\
    MixVPR & 4096 & 89.2 & 94.6 & 89.8 & 94.4 & 91.1 & 95.3 & 79.4 & 89.2 \\
    BoQ & 12288 & 91.5 & 96.5 & 91.1 & 95.7 & 93.1 & 96.6 & 83.8 & 92.2 \\
    \midrule
    EDTformer & 4096 & \textbf{92.0} & \textbf{96.6} & \textbf{92.4} & 95.9 & \textbf{93.4} & \textbf{97.0} & \textbf{88.3} & \textbf{95.3} \\
    EDTformer & 2048 & 91.4 & \textbf{96.6} & 90.6 & 95.7 & \textbf{93.4} & 96.7 & 82.3 & 91.9 \\
    EDTformer & 1024 & 91.4 & 95.7 & 90.6 & \textbf{96.0} & 92.5 & 96.6 & 78.6 & 89.8 \\
    EDTformer & 512 & 90.1 & 95.7 & 89.3 & 94.9 & 91.7 & 96.0 & 71.1 & 84.6 \\
    \bottomrule
    \end{tabular}
\end{table}

\textbf{Effect of EDTformer.} 
To validate the effectiveness of the proposed EDTformer, we compare it with current common aggregation methods, including GeM \cite{GeM}, NetVLAD \cite{netvlad}, Conv-AP \cite{gsv_cities}, MixVPR \cite{mixvpr} and BoQ \cite{BoQ} (related to our method). The ``GeM-linear'' represents the use of the GeM aggregator, followed by a linear layer to adjust the dimensionality of the descriptors. Likewise, ``NetVLAD-PCA'' implies that a PCA layer is integrated after the NetVLAD aggregator to perform dimensionality reduction. To ensure the fairness of the experiments, we consistently use the frozen DINOv2-base as the backbone with LoPA (rank set to 4) for feature extraction. The results are presented in Table \ref{tab:different aggregation methods}. When the dimensionality of global descriptors is the same (4096-dim), our EDTformer achieves highly competitive performance and outperforms other common aggregators (e.g., GeM, NetVLAD, Conv-AP and MixVPR) by a large margin. Although BoQ achieves outstanding results, our method continues to improve the performance, especially on the Nordland dataset (obtaining an improvement of 4.5\% in R@1). This is due to the fact that our EDTformer uses a simpler decoder structure to achieve more consistent feature aggregation. Notably, the dimensionality of the global descriptors (12288-dim) produced by BoQ is more than three times that of our global representations (4096-dim), which indicates our EDTformer can achieve higher retrieval efficiency than BoQ. Moreover, we simultaneously present the recognition performance of our global descriptors with varying dimensions (achieved by only adjusting the last linear layer in our pipeline). As the dimensionality of the descriptors decreases, the overall performance gradually declines. However, our 512-dim descriptors can still achieve an impressive 90.1\% R@1 on MSLS-val, 89.3\% R@1 on SPED, 91.7\% R@1 on Pitts30k and 71.1\% R@1 on Nordland, surpassing the performance of the 768-dim descriptors output by GeM and remaining competitive against some SOTA methods such as SFRS \cite{SFRS}, CosPlace \cite{Cosplace} and EigenPlace \cite{eigenplaces}. Therefore, our EDTformer is also well-suited for some VPR scenarios that urgently require low-dimensional descriptors.

\begin{table}[t]
    \centering
    \caption{Ablation on Different Fine-tuning/Adaptation Methods. ``PartialTuning-2'' and ``PartialTuning-4'' Respectively Denote Only Fine-tuning the Last 2 and 4 Encoder Blocks of DINOv2. ``MCAdapter'' Is the Abbreviation of MultiConvAdapter. The Trainable Parameters Introduced by Different Methods Are Reported. We Use an NVIDIA A800 GPU to Measure the Training Memory Usage with Batch Size Set to 72.}
    \label{tab:ablation of different fine-tuning methods}
    \setlength{\tabcolsep}{0.3mm}
    \begin{tabular}{l|c|c|cc|cc|cc}
    \toprule
    \multirow{2}{*}{Method} & \multicolumn{1}{c|}{Params} & \multicolumn{1}{c|}{Memory} & \multicolumn{2}{c|}{MSLS-val} & \multicolumn{2}{c|}{SPED} & \multicolumn{2}{c}{Nordland} \\
    \cmidrule{4-9} 
    & (M) & (GB) & R@1 & R@5 & R@1 & R@5 & R@1 & R@5 \\
    \midrule
    FrozenDINOv2 \cite{anyloc} & 0 & 8.35 & 90.8 & 96.1 & 90.1 & 95.6 & 74.3 & 86.3 \\
    \midrule
    PartialTuning-2 \cite{BoQ} & 14.18 & 15.54{\tiny \textcolor{red}{(+7.19)}} & 90.8 & 95.1 & 89.0 & 95.4 & 79.3 & 89.6 \\
    PartialTuning-4 \cite{SALAD} & 28.36 & 23.63{\tiny \textcolor{red}{(+15.28)}} & 91.1 & 95.4 & 86.0 & 92.9 & 74.1 & 85.5 \\
    FullTuning & 86.58 & 54.40{\tiny \textcolor{red}{(+46.05)}} & 88.0 & 95.4 & 83.0 & 91.6 & 72.3 & 84.7 \\
    Adapter \cite{selavpr} & 14.18 & 40.07{\tiny \textcolor{red}{(+31.72)}} & 91.1 & 96.2 & 89.6 & 94.6 & \textbf{88.8} & \textbf{95.6} \\
    MCAdapter \cite{cricavpr} & 9.16 & 39.30{\tiny \textcolor{red}{(+30.95)}} & 91.8 & 96.4 & 90.9 & 95.1 & 86.5 & 93.8 \\
    LoPA (ours) & \textbf{0.08} & \textbf{10.09}{\tiny \textcolor{red}{(+1.74)}} & \textbf{92.0} & \textbf{96.6} & \textbf{92.4} & \textbf{95.9} & 88.3 & 95.3 \\
    \bottomrule
    \end{tabular}
\end{table}
\textbf{Effect of LoPA.}
Next, to verify the efficiency and effectiveness of the proposed LoPA, we compare it with other common fine-tuning/PETL methods in VPR, including PartialTuning used in BoQ \cite{SALAD} and SALAD \cite{SALAD} (only fine-tuning the last few encoder blocks of the backbone), FullTuning (training the whole backbone), Adapter (as implemented in SelaVPR \cite{selavpr}) and MultiConvAdapter (as implemented in CricaVPR \cite{cricavpr}). We use DINOv2-base as the backbone and set the frozen backbone without any adjustment as baseline. Besides, we consistently use our EDTformer as the aggregator. The results are presented in Table \ref{tab:ablation of different fine-tuning methods}. 
\textbf{(1)} In terms of efficiency, LoPA is an extremely lightweight network, solely introducing 0.08M trainable parameters, fewer than 1\% of the parameters compared to other methods. In addition, compared to the frozen baseline, it only results in 1.74GB additional training memory usage, approximately 1/4 of the PartialTuning-2 method and even 10 times less than the PETL methods used in SelaVPR and CricaVPR. Amazingly, it saves over 40GB training memory usage compared to FullTuning. This demonstrates that LoPA is parameter- and memory-efficient, which makes it highly appropriate for VPR scenarios with constrained computational resources. 
\textbf{(2)} In terms of accuracy, LoPA also significantly improves the performance of DINOv2 in VPR, while other methods result in minimal improvements or even a decline. Specifically, without any adjustment to DINOv2, our EDTformer remains competitive against some SOTA methods, such as MixVPR \cite{mixvpr} and EigenPlaces \cite{eigenplaces}. Building on this, our LoPA further improves the baseline with 1.2\%, 14.0\% and 2.3\% absolute R@1 on MSLS-val, Nordland and SPED. In contrary, PartialTuning yields minimal performance improvement on MSLS-val and Nordland. FullTuning, on the other hand, results in an overall performance decline. Although the PETL methods, i.e., Adapter and MultiConvAdapter, improve the recognition performance of the model on the MSLS-val and Nordland, they lead to a decrease in R@5 on SPED. Comprehensively, the features output by DINOv2 with LoPA are more suitable for the sequent EDTformer to form global representations, which is one of the reasons why our model performs better than other methods (e.g., BoQ, SALAD, etc).

\begin{table}[t]
    \centering
    \setlength{\tabcolsep}{1.15mm}
    \caption{Ablation on the Number of Decoder Blocks. The Parameters of Decoder Blocks Are Reported. We Consistently Use DINOv2-base as the Backbone with LoPA (Rank Set to 4) and Leverage 64 Queries as One of the Inputs for EDTformer.}
    \label{tab:ablation on the number of decoder layers}
    \begin{tabular}{c|c|c c | c c | c c | c c}
    \toprule
    \multirow{2}{*}{Number} & \multicolumn{1}{c|}{Params} & \multicolumn{2}{c|}{MSLS-val} & \multicolumn{2}{c|}{SPED} & \multicolumn{2}{c|}{Pitts30k} & \multicolumn{2}{c}{Nordland} \\
    \cmidrule{3-4} \cmidrule{5-6} \cmidrule{7-8} \cmidrule{9-10}
    & (M) & R@1 & R@5 & R@1 & R@5 & R@1 & R@5 & R@1 & R@5 \\
    \midrule
    1 & 4.73 & 91.8 & 96.6 & 92.1 & 96.0 & 93.1 & 96.6 & 78.4 & 88.4 \\
    2 & 9.46 & 92.0 & 96.6 & \textbf{92.4} & 95.9 & \textbf{93.4} & \textbf{97.0} & \textbf{88.3} & \textbf{95.3} \\
    3 & 14.18 & \textbf{92.6} & 96.2 & 91.3 & \textbf{96.4} & 93.0 & 96.7 & 86.9 & 94.6 \\
    4 & 18.91 & 91.9 & \textbf{96.9} & 90.9 & 95.9 & 93.1 & 96.9 & 88.2 & \textbf{95.3} \\
    6 & 28.37 & 91.8 & 96.2 & 89.3 & 94.4 & 92.7 & 96.5 & 84.0 & 92.9 \\
    \bottomrule
    \end{tabular}
\end{table}

\begin{table}[t]
    \centering
    \setlength{\tabcolsep}{1.2mm}
    \caption{Ablation on the Number of Learnable Queries. We Consistently Use DINOv2-base as the Backbone with LoPA (Rank Set to 4). The Overall Performance Is Best with 64 Queries.}
    \label{tab:ablation_of_number_queries}
    \begin{tabular}{c | c c | c c | c c | c c}
    \toprule
    \multirow{2}{*}{\makecell{Number of \\ learnable queries}} & \multicolumn{2}{c|}{MSLS-val} & \multicolumn{2}{c|}{SPED} & \multicolumn{2}{c|}{Pitts30k} & \multicolumn{2}{c}{Nordland} \\
    \cmidrule{2-3} \cmidrule{4-5} \cmidrule{6-7} \cmidrule{8-9}
    & R@1 & R@5 & R@1 & R@5 & R@1 & R@5 & R@1 & R@5 \\ 
    \midrule
    8 & 91.1 & 96.5 & 91.4 & 95.9 & 92.8 & 96.7 & 77.4 & 88.5 \\
    16 & 92.2 & 96.5 & 90.8 & 95.9 & 92.7 & 96.7 & 80.9 & 91.0 \\
    32 & 91.5 & 96.2 & 91.4 & 96.0 & 93.2 & 96.7 & 86.1 & 94.3 \\
    64 & 92.0 & \textbf{96.6} & \textbf{92.4} & 95.9 & \textbf{93.4} & \textbf{97.0} & \textbf{88.3} & \textbf{95.3} \\
    96 & \textbf{92.6} & 96.5 & 92.3 & \textbf{96.4} & 93.0 & \textbf{97.0} & 85.4 & 94.0 \\
    \bottomrule
    \end{tabular}
\end{table}

\textbf{Effect of the number of decoder blocks.}
To evaluate the impact of the number of decoder blocks used in our EDTformer architecture, we conduct the ablation experiments by changing the number of decoder blocks. The results, as shown in \ref{tab:ablation on the number of decoder layers}, demonstrate that even using one decoder block, EDTformer can still get a competitive performance compared to some SOTA methods, such as MixVPR \cite{mixvpr}, CricaVPR \cite{cricavpr} and SelaVPR \cite{selavpr}. The overall best performance is achieved by constructing the EDTformer with two stacked decoder blocks, which can fully decode and aggregate the crucial contextual information from the deep features. In contrast, further increasing the number of decoders (e.g., 3 and 4) provides very limited improvement and even slightly degrades the overall performance. Besides, there is an obvious drop in recognition accuracy when the number of decoder blocks increases to 6. Moreover, the increase in parameters and memory consumption caused by stacking more decoder blocks is also a non-negligible issue. For a better trade-off between efficiency and accuracy, using 2 decoder blocks to construct EDTformer is a good choice.

\textbf{Effect of the number of learnable queries.}
In this part, we conduct the ablation studies on the number of learnable queries. The results are presented in Table \ref{tab:ablation_of_number_queries}. Since the learnable queries are an essential input of EDTformer, their quantity affects the quality of the global features. We can observe that the overall performance of the model improves as the number of learnable queries increases. However, when the number of queries becomes too large, it not only introduces an additional computational burden but also is unable to sufficiently learn effective features, which hinders the generation of a robust and discriminative global representation. In other words, few queries may only aggregate limited information, while too many queries may lead to information redundancy. Both cases are unlikely to achieve the best results. To achieve a trade-off between accuracy and efficiency, we employ 64 queries for optimal overall performance.

\begin{table}[t]
    \centering
    \caption{Ablation on FFN in the Decoder Block. The Total Trainable Parameters, R@1 and R@5 Performance Are Reported.}
    \label{tab:Ablation on the FFN}
    \setlength{\tabcolsep}{0.8mm}
    \begin{tabular}{c|c|c|cc|cc|cc|cc}
    \toprule
    \multirow{2}{*}{FFN} & \multirow{2}{*}{LoPA} & \multicolumn{1}{c|}{Params} & \multicolumn{2}{c|}{MSLS-val} &  \multicolumn{2}{c|}{SPED} & \multicolumn{2}{c|}{Pitts30k} & \multicolumn{2}{c}{Nordland}  \\
    \cmidrule{4-11} 
    & & (M) & R@1 & R@5 & R@1 & R@5 & R@1 & R@5 & R@1 & R@5 \\
    \midrule
    \checkmark & \multirow{2}{*}{$\times$} & 16.59 & 90.5 & \textbf{96.2} & 89.6 & 95.1 & 91.6 & \textbf{96.4} & \textbf{74.9} & \textbf{86.9} \\
    $\times$ & & 10.29 & \textbf{90.8} & 96.1 & \textbf{90.1} & \textbf{95.6} & \textbf{91.8} & 96.3 & 74.3 & 86.3 \\
    \midrule 
    \checkmark & \multirow{2}{*}{\checkmark} & 16.68 & 91.9 & \textbf{96.6} & 90.8 & \textbf{96.0} & 92.8 & 96.6 & 84.2 & 92.7 \\
    $\times$ &  & 10.38 & \textbf{92.0} & \textbf{96.6} & \textbf{92.4} & 95.9 & \textbf{93.4} & \textbf{97.0} & \textbf{88.3} & \textbf{95.3} \\
    \bottomrule
    \end{tabular}
\end{table}

\begin{table}[t]
    \centering
    \caption{Ablation on the Rank of LoPA. We Extensively Explore How Different Ranks Impact the Performance. DINOv2-base Is Used as the Backbone. ``w/o LoPA'' Means Only Using Frozen DINOv2 Without Any Fine-tuning. The Overall Performance Is Best with Rank Set to 4.}
    \label{tab:Ablation of the rank of LoPA}
    \setlength{\tabcolsep}{1.6mm}
    \begin{tabular}{c|c c | c c | c c | c c}
    \toprule
    \multirow{2}{*}{Rank} & \multicolumn{2}{c|}{MSLS-val} & \multicolumn{2}{c|}{Nordland} & \multicolumn{2}{c|}{Pitts30k} & \multicolumn{2}{c}{SPED}  \\
    \cmidrule{2-3} \cmidrule{4-5} \cmidrule{6-7} \cmidrule{8-9} 
    & R@1 & R@5 & R@1 & R@5 & R@1 & R@5 & R@1 & R@5 \\
    \midrule
    w/o LoPA & 90.8 & 96.1 & 74.3 & 86.3 & 91.8 & 96.3 & 90.1 & 95.6 \\ 
    \midrule
    2 & 91.9 & \textbf{96.6} & 86.4 & 94.2 & 92.9 & 96.5 & \textbf{92.8} & 96.5 \\
    4 & 92.0 & \textbf{96.6} & \textbf{88.3} & \textbf{95.3} & \textbf{93.4} & \textbf{97.0} & 92.4 & 95.9 \\
    8 & 92.3 & 96.2 & 87.3 & 94.9 & 92.8 & 96.6 & 91.4 & 96.0 \\
    16 & 91.9 & \textbf{96.6} & 84.2 & 92.9 & 93.2 & 96.9 & 91.3 & 96.0 \\
    24 & \textbf{92.7} & 95.9 & 87.7 & 94.7 & 93.2 & 96.7 & 91.8 & \textbf{96.9} \\
    \bottomrule
    \end{tabular}
\end{table}

\begin{table}[t]
    \centering
    \caption{Ablation of LoPA on Other VPR Methods. We Consistently Use the Frozen DINOv2-base Backbone.}
    \label{tab:Other_VPR_methods_LoPA}
    \setlength{\tabcolsep}{0.5mm}
    \begin{tabular}{l|c|c c|c c|c c|c c}
    \toprule
    \multirow{2}{*}{Method} & \multirow{2}{*}{LoPA} & \multicolumn{2}{c|}{Pitts30k} & \multicolumn{2}{c|}{MSLS-val} & \multicolumn{2}{c|}{Nordland} & \multicolumn{2}{c}{SPED} \\
    \cmidrule{3-10}
    & & R@1 & R@5 & R@1 & R@5 & R@1 & R@5 & R@1 & R@5 \\
    \midrule
    \multirow{2}{*}{GeM} & $\times$ & 78.4 & 89.2 & 36.4 & 47.0 & 12.1 & 21.3 & 58.5 & 75.3 \\
    & \checkmark & \textbf{88.5} & \textbf{94.4} & \textbf{85.7} & \textbf{94.6} & \textbf{32.3} & \textbf{47.1} & \textbf{76.1} & \textbf{88.5} \\
    \midrule
    \multirow{2}{*}{NetVLAD} & $\times$ & 82.7 & 92.8 & 55.3 & 67.7 & 31.1 & 46.2 & 80.2 & 92.3 \\
    & \checkmark & \textbf{92.0} & \textbf{96.3} & \textbf{89.7} & \textbf{95.5} & \textbf{71.7} & \textbf{84.8} & \textbf{88.8} & \textbf{95.1} \\
    \midrule
    \multirow{2}{*}{SALAD} & $\times$ & 91.5 & 96.1 & 88.9 & 95.4 & 62.0 & 75.5 & 88.3 & 94.4 \\
    & \checkmark & \textbf{92.5} & \textbf{96.4} & \textbf{91.6} & \textbf{96.2} & \textbf{84.1} & \textbf{92.5} & \textbf{91.3} & \textbf{96.2} \\
    \bottomrule
    \end{tabular}
\end{table}

\begin{figure*}[ht]
    \centering
    \vspace{-0.4cm}
    \includegraphics[width=0.88\textwidth]{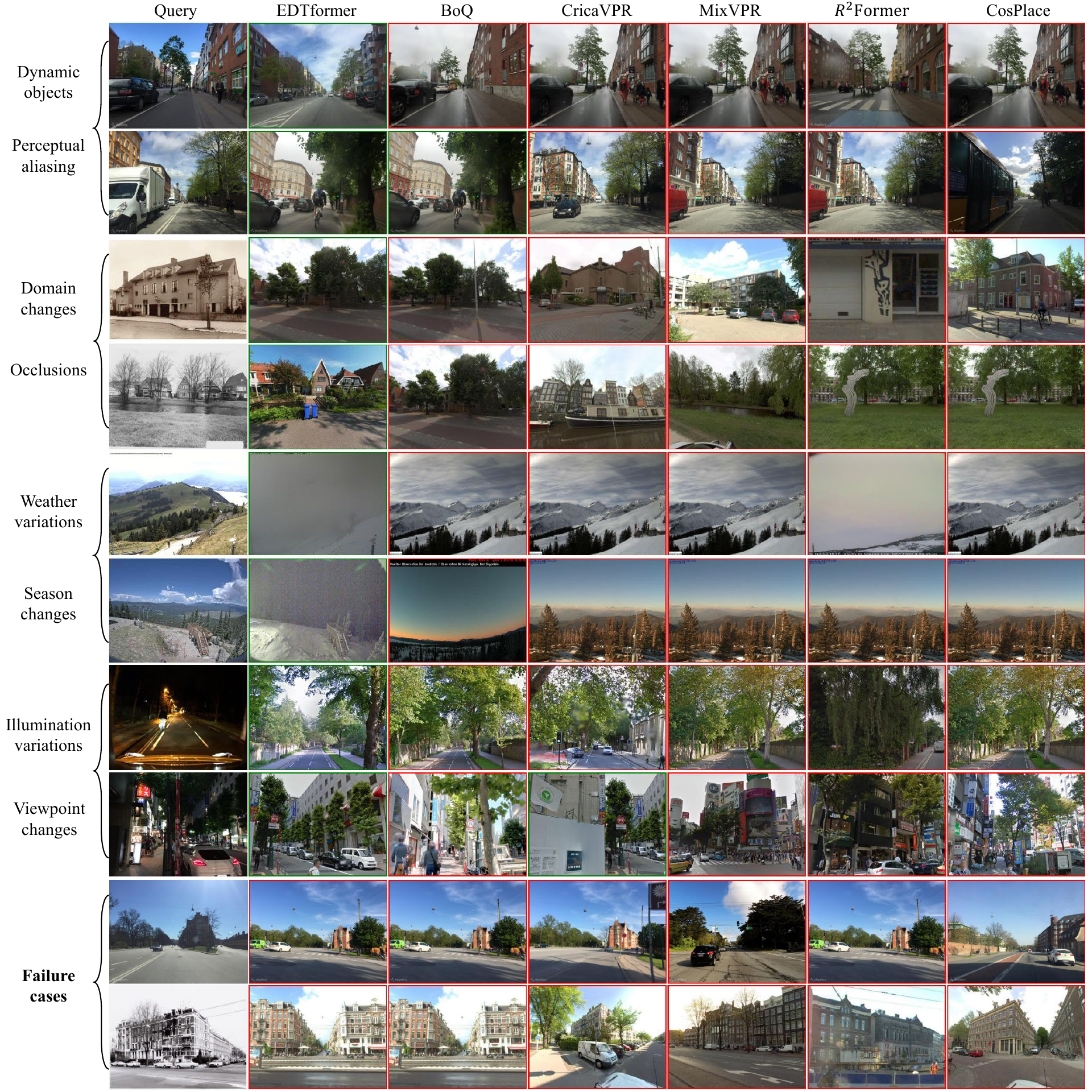}
    \caption{Qualitative results. In these challenging scenarios, our method successfully retrieves the correct images, while other methods commonly return the false places.
    For the first two examples, although some other methods obtain images geographically close to the query image, they exceed the set threshold (25m).
    For the third and fourth examples, despite image modality changes between the query and database images, our method still can retrieve the correct places by capturing the invariant and discriminative buildings.
    For the fifth and sixth examples, the query images are captured in natural scenes, suffering from severe condition variations and lacking discriminative landmarks. Nevertheless, our method can still match the correct place.
    In the seventh and eighth examples, other methods commonly return a false result due to the severe lighting changes. However, our method can produce robust and discriminative global descriptors, which can effectively handle the problem. 
    For the last two examples, all methods fail when facing extremely difficult scenarios, in which viewpoint changes, domain variations, occlusions, dynamic objects and perceptual aliasing arise simultaneously.
    }
    \label{fig:matching}
    \vspace{-0.3cm}
\end{figure*}

\begin{figure*}[!htbp]
    \centering
    \vspace{-0.5cm}
    \includegraphics[width=1.0\textwidth]{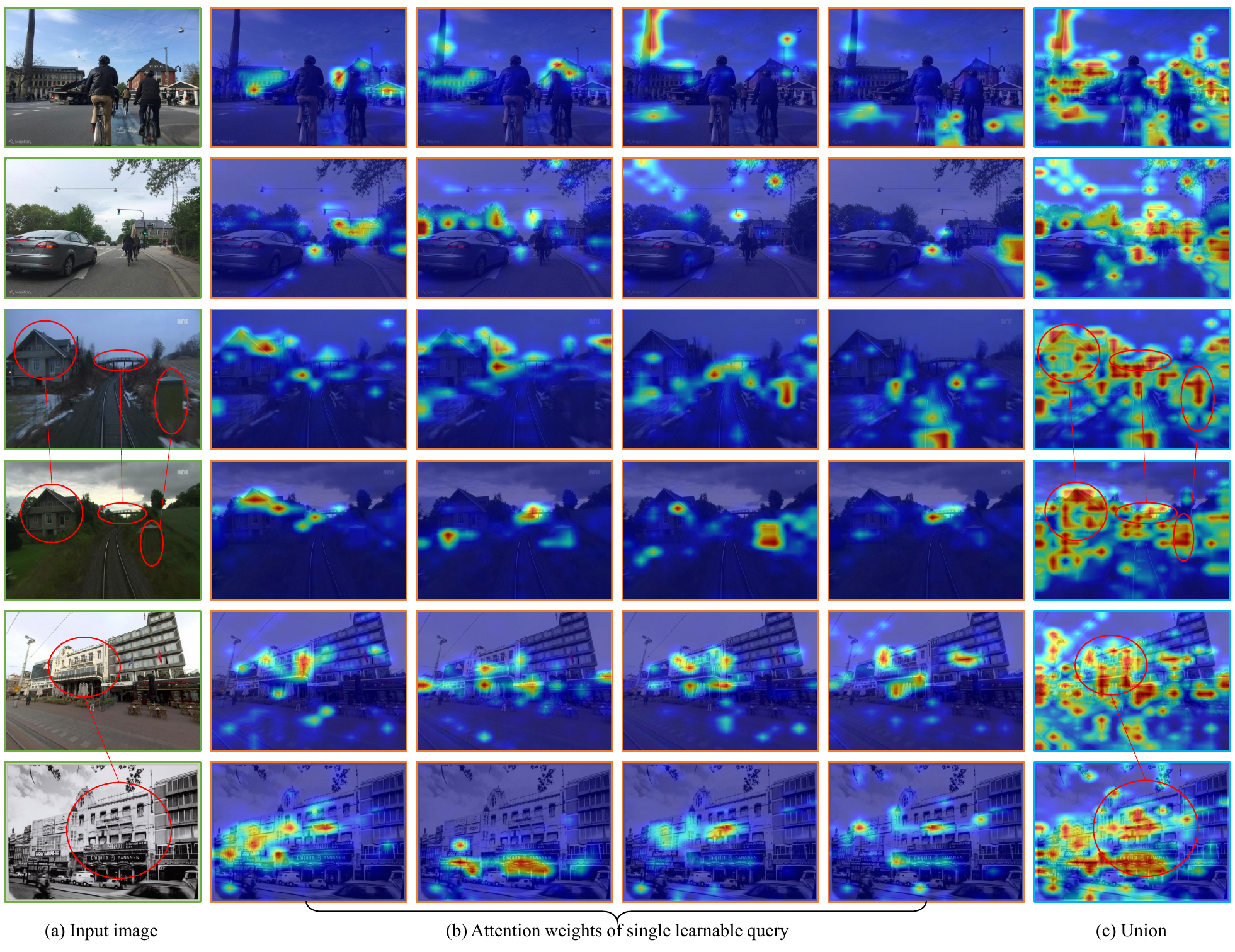}
    \caption{Visualization of learnable queries. The first column (a) with a green border shows the original input images, the middle 2-5 columns (b) with an orange border represent the attention weights of a single learnable query, and the last column (c) with a blue border denotes the union of all query attention weights. It is worth noting that different queries prioritize different areas in the images, but they all focus on the temporarily constant and discriminative regions. On the contrary, for variable elements, these queries tend to discard them. Concretely, in the first two examples, all queries pay attention to the discriminative and short-term unchanged buildings, trees, and special signs on the ground while dynamic pedestrians and vehicles are basically ignored. The third and fourth examples are taken from almost the same location but with different illumination and seasons. Nevertheless, our learnable queries consistently focus on the condition-invariant house, bridge, and special landmark. When the place images (i.e., the last two examples) come from different perspectives and span a very long time, our method can pay attention to the discriminative and unchanged regions that they commonly possess.}
    \vspace{-0.4cm}
    \label{fig:visualization}
\end{figure*}

\textbf{Effect of the FFN in the decoder block.}
We simplify the standard transformer decoder block by removing the feedforward network (FFN) to build our EDTformer for higher efficiency. Here, we conduct the corresponding experiments to explain why we do that. We consistently utilize two decoder blocks to construct EDTformer with 64 learnable queries. For comprehensiveness, we conduct the experiments with and without LoPA. The results are presented in Table \ref{tab:Ablation on the FFN}. Without LoPA to adapt the frozen DINOv2, our simplified decoder performs similarly to the standard transformer decoder. With our LoPA, however, using the simplified decoder can achieve a better overall recognition performance, such as an absolute 4.1\% R@1 improvement on Nordland. This indicates the features produced by DINOv2 with LoPA are more suitable for the sequent EDTformer consisting of our simplified decoder blocks. Besides, our simplified decoder introduces less trainable parameters and can achieve higher efficiency compared to the standard transformer decoder. Therefore, using our simplified decoder block is a better choice, which is advantageous without any drawbacks.

\textbf{Effect of the rank of LoPA.}
In this subsection, we further conduct ablation studies about the effect of the rank in LoPA. We consistently use the EDTformer for feature aggregation. Table \ref{tab:Ablation of the rank of LoPA} shows the results of setting different ranks. We set ``w/o LoPA'' as baseline. Setting the rank to 4 gets the best overall performance, but we can still achieve good results with rank even set to 2. These results show that we can achieve an outstanding performance without introducing a large number of parameters and consuming substantial memory during training. It is worth mentioning that the backbone DINOv2 is completely frozen, which indicates that the intermediate features produced by each block of DINOv2 remain consistent. The final output solely depends on how these intermediate activations are processed, which enlightens our future work to design a better parallel network to refine them. Meanwhile, due to only introducing a few parameters and memory usage, it can facilitate the application of vision foundation models under resource-limited conditions.

\textbf{Effect of LoPA on other VPR methods.}
To further demonstrate the universality of our proposed LoPA, we conduct ablation studies about the effect of LoPA on some other common VPR methods, including GeM, NetVLAD, and SALAD. The results are reported in Table \ref{tab:Other_VPR_methods_LoPA}. Only using the DINOv2-base backbone without LoPA to adapt it for the VPR task leads to poor performance. However, with the help of LoPA, all these VPR methods can achieve a significant improvement. For example, GeM with LoPA achieves nearly 3$\times$ higher R@1 on Nordland compared to its baseline. These results indicate that our proposed LoPA is not tailored for our EDTformer, but also is applicable to some other VPR methods.

\subsection{Qualitative Experiments}
Fig. \ref{fig:matching} presents the qualitative experimental results in various challenging environments, including viewpoint variants, drastic condition changes, severe occlusions, image domain variations, etc. In the vast majority of cases, our method successfully gets similar and correct place images from the same location as the query images, while other methods tend to struggle with obtaining the correct images within the predefined threshold (25m), which demonstrates that our method is highly robust to environment and viewpoint changes, as well as less prone to perceptual aliasing. For instance, in the seventh and eighth examples, the query images are captured at night. Only a small region of the image is clearly visible. However, our method still gets the correct results. Additionally, it is worth mentioning that all methods fail to obtain the correct places in the last two examples, where multiple challenges (e.g., viewpoint changes, domain variations, severe occlusions, dynamic objects and perceptual aliasing) arise simultaneously. This motivates us to further improve our method in future work to get higher accuracy and tackle more challenging scenarios.

\subsection{Interpretability Analysis}
The experimental results presented in Table \ref{tab:comparison with state-of-the-art} and \ref{tab:Performance in extreme scenarios} have fully demonstrated the superior performance of our method. In this subsection, we further explore the underlying reasons. To this end, we visualize the attention weights between the input image and learnable queries within our EDTformer to understand their unique aggregating characteristics. The results are shown in Fig. \ref{fig:visualization}. Each single learnable query has a different focus, but they all share a common characteristic. Observing horizontally (especially the first two examples), we can find that each learnable query basically focuses on short-term invariant yet discriminative regions (e.g., buildings, trees, and some special signs on the ground) while ignoring dynamic and easily changing objects (e.g., sky, vehicles, and pedestrians) that are useless for the VPR task \cite{onlylookonce,selavpr}. Moreover, when facing various challenges (e.g., illumination and seasonal changes, viewpoint changes, etc) as shown in the last four examples, our method can also tackle them well. These results indicate that our EDTformer successfully decodes crucial features for the VPR task into these learnable queries to achieve feature aggregation. With all information within the learnable queries combined, they can consistently concentrate on the majority of discriminative regions in the place images, which is why our method can achieve a great recognition performance and effectively address most challenges in VPR.

\section{Conclusions}
In this paper, we revisited the transformer decoder and proposed a novel feature aggregation method EDTformer for VPR, which is simple, efficient and powerful. The EDTformer, primarily consisting of our simplified decoder blocks, can fully decode and aggregate the effective information within deep features extracted from the backbone into a set of learnable queries, thus producing a robust and discriminative global representation for the place image. Moreover, we designed a Low-rank Parallel Adaptation method, which can adapt the foundation model DINOv2 for the VPR task in a memory- and parameter-efficient way, thereby providing more powerful features for our EDTformer and enhancing the robustness of the entire model. The extensive experiments show that our approach surpasses other SOTA methods on multiple benchmark datasets by a large margin with less training memory usage and effectively tackles various challenges in VPR.

\bibliographystyle{IEEEtran}
\bibliography{IEEEabrv,main}

\end{document}